\documentclass[10pt,twocolumn,letterpaper]{article}

\usepackage{iccv}
\usepackage{times}
\usepackage{epsfig}
\usepackage{graphicx}
\usepackage{amsmath}
\usepackage{amssymb}
\usepackage{caption}
\usepackage{algorithm}
\usepackage{algorithmic}
\usepackage{booktabs}
\usepackage{makecell}
\usepackage{textcomp}

\usepackage[accsupp]{axessibility} % Improves PDF readability for those with disabilities.

\usepackage[pagebackref=true,breaklinks=true,letterpaper=true,colorlinks,bookmarks=false]{hyperref}
\usepackage{color,soul}

\usepackage{etoolbox}
\patchcmd{\thebibliography}{\section*{\refname}}{}{}{}

\iccvfinalcopy % *** Uncomment this line for the final submission

\def\iccvPaperID{****} % *** Enter the ICCV Paper ID here

\begin{document}

%%%%%%%%% TITLE
\title{ProAI: An Efficient Embedded AI Hardware for Automotive Applications\\ – a Benchmark Study}

\author{Sven Mantowsky\textsuperscript{1}, Falk Heuer\textsuperscript{1}, Syed Saqib Bukhari\textsuperscript{1}, Michael Keckeisen\textsuperscript{2}, Georg Schneider\textsuperscript{1}\\
{\footnotesize \textsuperscript{1}ZF Friedrichshafen AG, Artificial Intelligence Lab, Saarbrücken, Germany  \textsuperscript{2}ZF Friedrichshafen AG, Autonomous Mobility Systems, Germany}}

\maketitle
\ificcvfinal\thispagestyle{empty}\fi

%%%%%%%%% ABSTRACT
\begin{abstract}
  Development in the field of Single Board Computers (SBC) have been increasing for several years. They provide a good balance between computing performance and power consumption which is usually required for mobile platforms, like application in vehicles for Advanced Driver Assistance Systems (ADAS) and Autonomous Driving (AD). However, there is an ever-increasing need of more powerful and efficient SBCs which can run power intensive Deep Neural Networks (DNNs) in real-time and can also satisfy necessary functional safety requirements such as Automotive Safety Integrity Level (ASIL). “ProAI” is being developed by ZF mainly to run powerful and efficient applications such as multitask DNNs and on top of that it also has the required safety certification for AD. 
  In this work, we compare and discuss state of the art SBC on the basis of power intensive multitask DNN architecture called Multitask-CenterNet with respect to performance measures such as, FPS and power efficiency. As an automotive supercomputer, ProAI delivers an excellent combination of performance and efficiency, managing nearly twice the number of FPS per watt than a modern workstation laptop and almost four times compared to the Jetson Nano. Furthermore, it was also shown that there is still power in reserve for further and more complex tasks on the ProAI, based on the CPU/GPU utilization during the benchmark.

\end{abstract}

%%%%%%%%% BODY TEXT
\section{Introduction}
\label{sec:intor}
The progress in the field of Deep Neural Networks (DNNs) have increased exponentially since the advent of AlexNet \cite{alxnet2012} back in 2012. In last decade significant progress has been made in several domains such as natural language processing \cite{trendslanguageprocess2017} and computer vision \cite{deepreinforcement2017}\cite{surversemseg2018}. Their applications are already established in different areas; strongly including the automotive sector like Advanced Driver Assistance Systems (ADAS) and Autonomous Driving (AD). For the development of deep learning-based methods Graphics Processing Units (GPUs) are mainly used, which have enormously large parallel computing power. Furthermore, other processing units are also developed for these methods such as Tensor Processor Units (TPU~\footnote{https://cloud.google.com/tpu/docs/tpus}) or even entire workstations that make parallel use of GPUs possible (such as Nvidia DGX-2~\footnote{https://www.nvidia.com/de-de/data-center/dgx-2/}). However, the high computing power also requires a lot of electrical power that cannot be provided under mobile circumstances in vehicles.

\begin{figure}[t]
\begin{center}
\includegraphics[width=0.5\textwidth]{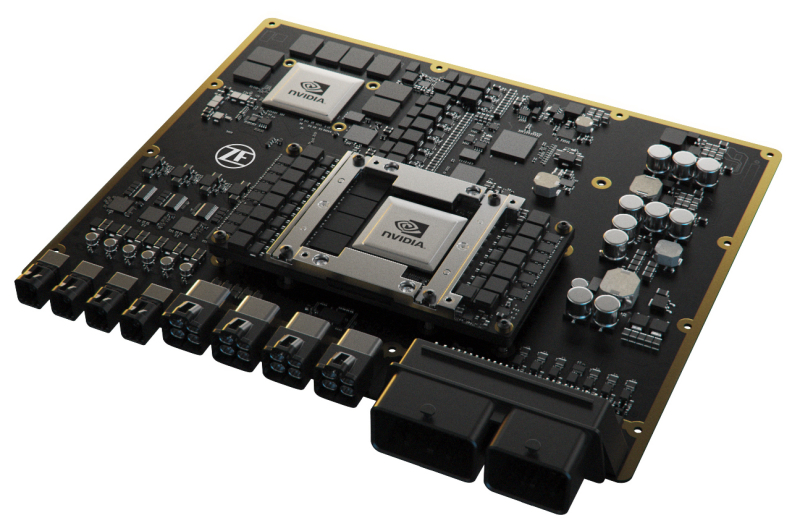}
\end{center}
   \caption{Example of ZF’s Single Board Computer: the ProAI.}
\label{fig:proai}
\end{figure}

Development in the field of Single Board Computers (SBC) has been increasing for several years for achieving the best possible balance between computing performance and power consumption mainly for mobile platforms. There are many manufacturers who produce chips for these boards, such as Texas Instruments, Xilinx, Qualcomm and NVIDIA. The boards with GPU support are providing computing power that is required for mobile platforms, because they combine high core density of the processors with lower energy consumption. However, there is a persistent need of even more powerful and efficient Single Board Computers which on one side can run multiple Deep Neural Networks (DNNs) together in real-time, and on the other side, can satisfy necessary functional safety requirements for autonomous driving such as Automotive Safety Integrity Level (ASIL) \cite{iso26262}. 

ProAI is being developed by ZF mainly to solve these issues. It is specifically designed for running powerful and efficient applications such as multitask DNNs \cite{objectsaspoints2019,multitask1997,maskrcnn2017,panoptic2018} in real-time for autonomous driving. On top of that, ProAI already has ASIL certification. An example of ProAI board is shown in Figure~\ref{fig:proai}. For benchmarking ProAI and comparison with other hardware, the Multitask-CenterNet architecture \cite{efficientmultitask2021} is used here. It is a power intensive architecture, which can perform multiple perception related tasks such as semantic segmentation, object detection, and human pose estimation in parallel. It can be used with different configurations that range from single to multiple tasks in parallel.

This paper provides the following main contributions: i) Presentation of the mobile single board computer “ProAI” \cite{proai} for DNNs. ii) Benchmark result for the ProAI are presented for the Multitask-CenterNet architecture. For this benchmark, power, latency, and frames per second (FPS) metrics are addressed which are suitable to judge the performance of mobile platforms. iii) Comparison of ProAI with different platforms including the NVIDIA Jetson Nano (as another mobile platform) and a typical workstation laptop. Altogether in this paper we show on one hand the advantage of a modern multitasks network architecture such as the Multitask-CenterNet for mobile application in vehicles and on the other hand the potential of ZF's ProAI as compared to other platforms for such power intensive DNN architectures. 

The paper is further organized as follows: In Section~\ref{sec:proai} we present technical specifications of ProAI in comparison to different mobile platforms. In Section~\ref{sec:mcenternet} the architecture of the Multitask-CenterNet is described. Section~\ref{sec:dataset} highlights the dataset, metrics and benchmark method used in this paper. In Section~\ref{sec:benchmark} the benchmark on different platforms are shown and discussed. Finally the paper is concluded in Section~\ref{sec:conclusion}.

%------------------------------------------------------------------------
\section{ProAI – Single Board Computer with GPU}
\label{sec:proai}

%...

\begin{table}[h!]
\fontsize{8}{9}\selectfont
  \begin{center}
    
    \begin{tabular}{l|c|c|c}
      \toprule % <-- Toprule here
      \textbf{} & \bfseries\makecell{Jetson Nano} & \bfseries\makecell{Dell Precision \\ 7530} & \textbf{ProAI}\\

      \midrule % <-- Midrule here
      \midrule
      \bfseries\makecell{Perform\\\big[GFLOPS\big]}  & \makecell{472$@$FP16\\235$@$FP32\\7.3$@$FP64} & \makecell{38.5$@$FP16\\2468$@$FP32\\77.14$@$FP64} & \makecell{2820$@$FP16\\1410$@$FP32\\705$@$FP64}\\
      \midrule
      \bfseries\makecell{CPU} & \makecell{Quad-core\\ARM Cortex-A72\\64-bit$@$\\1.5GHz} & \makecell{Intel\textsuperscript{\textregistered} Core\textsuperscript{\texttrademark} i7-\\8850H, 6 Cores$@$\\2.6GHz} & \makecell{Octa-core\\ARMv8 64-bit}\\
      \midrule
      \bfseries\makecell{GPU} & \makecell{Maxwell 2.0\\GM20B} & Pascal GP107 & Volta GV10B\\
      \midrule
      \bfseries\makecell{GPU-\\Memory} & \makecell{4GB LPDDR4\\$@$1600MHz\\25.6GB/s} & \makecell{4GB GDDR5\\$@$1502MHz\\96.13GB/s} & \makecell{32GB LPDDR4\\$@$1600MHz\\127GB/s}\\
      \midrule
      \bfseries\makecell{GPU-\\Clock\\(boost)} & 921 MHz & 1557 MHz & 1377 MHz\\
      \midrule
      \bfseries\makecell{GPU-\\Cores} & 128 Cores & 768 Cores & \makecell{512 Cores \&\\64 Tensor Cores}\\
      \midrule
      \bfseries\makecell{CUDA} & .52 & 6.1 & 7.2\\
      \midrule
      \bfseries\makecell{Internal\\Memory}  & SDHC slot & \makecell{32GB DDR4 RAM\\1TB HDD} & \makecell{32GB \\eMMC 5.1}\\
      \midrule
      \bfseries\makecell{Video\\Encode}  & \makecell{1x4Kp30\\2x1080p60\\4x1080p30\\4x720p60\\9x720p30\\(H.265\&\\H.264)} & \makecell{4x4Kp120\\4x5Kp60\\(H.264\&HEVC)} & \makecell{4x4Kp60\\8x4Kp30\\16x1080p60\\32x1080p30\\(H.265)\\4x4Kp60\\8x4Kp30\\14x1080p60\\30x1080p30\\(H.264)}\\
      \midrule
      \bfseries\makecell{Video\\Decode}  & \makecell{1x4Kp60\\1x4Kp30\\4x1080p60\\8x1080p30\\9x720p60\\(H.265\&\\H.264)} & \makecell{4x4Kp60\\H.264\\HEVC\\MVC\\VC-1} & \makecell{2x8Kp30\\6x4Kp60\\12x4Kp30\\26x1080p60\\52x1080p30\\(H.265)\\4x4Kp60\\8x4Kp30\\16x1080p60\\32x1080p30\\(H.264)}\\
      \midrule
      \bfseries\makecell{Camera}  & MIPI CSI & USB & CAN, USB\\
      \midrule
      \bfseries\makecell{GPU\\Power\\(max)}  & 5-10W & 75W & 15-30W\\
      \midrule
      \bottomrule % <-- Bottomrule here
    \end{tabular}
    \caption{Comparison between different Platforms used in this work: NVIDIA Jetson Nano, Dell Precision 7530 Workstation and ZF's ProAI.}
    \label{fig:tableproai}
  \end{center}
\end{table}

This section first gives a technical overview of Single Board Computers and then further details about ZF’s ProAI.

\subsection{Single Board Computers (SBC)}
\label{subsec:sbc}
Single Board Computers form a complete system in themselves, just like laptop or desktop computers. In contrary to laptop or desktop computers, all of SBC components are installed on the same board. Their architecture is geared towards efficiency and therefore mostly built on ARM-based central processing chips. They cannot perform as many functions as a powerful desktop computer, but they represent a great compromise between energy consumption and computing power. These boards are available with a wide variety of processing units such as Central Processing Unit (CPU), Multiprocessing Unit (MPU), Graphics Processing Unit (GPU) or Field Programable Gate Array (FPGA) and several combinations thereof. Due to their compact design and their significantly lower power consumption, they are very well suited for mobile applications, such as robotics, drones, and vehicle technology. One of the most popular SBCs is the Raspberry PI~\footnote{https://www.raspberrypi.org/}. It offers the performance of a small laptop, for the size of a mobile phone.  It is well suited for multimedia, HD streaming or games, as well as the possibility of text and image processing in 4K resolution at a refresh rate of up to 60Hz. However, when it comes to deep learning, memory and neural network model sizes are the limitations. SBCs with focus on machine learning often have a dedicated GPU and therefore a clear advantage in terms of computing power. Due to the immense parallelization possibilities of the GPU’s, boards like the NVIDIA Jetson NANO offer a much better opportunity for the computational needs of deep and complex neural networks. Above all, the NVIDIA GPU architecture offers a great advantage over other manufacturers: the CUDA library. This opens up the possibility of calculating all arithmetic operations on the GPU. In a perception task like semantic segmentation are innumerable calculations during the inference procress. 
In practical applications, for example in automotive sector, there are usually additional requirements such as compatibility with the CAN-BUS, automotive-grade software (e.g. AUTOSAR), quality ensurance, safety and security. Many design guidelines, hardware and software requirements must be considered during the development process to obtain ASIL certification in accordance with the ISO 26262/AEC-Q100 standard \cite{iso26262}. This certification is necessary in European area, to allow vehicles with new electrical components to drive on public roads. Most of the existing boards do not fulfil these requirements.

\subsection{ProAI}
\label{subsec:proai}

ZF has taken up the current challenges of SBC and successfully developed an automotive-grade single board computer that both meets the quality and safety requirements, and furthermore provides the required computing power for a wide range of Deep Learning applications. Regarding the nature of the board which is suitable for ADAS and Deep Learning based tasks it has been given the name “ProAI”. In addition to a very competent octa-core CPU (ARM-architecture), a powerful NVIDIA graphics processor provides the ''computational heart'' of ProAI. Some essential parameters are listed in Table ~\ref{fig:tableproai} and compared with other GPU-based platforms like NVIDIA’s Jetson Nano and the workstation used in this work. Based on the parameter values, the potential of ProAI can already be projected, which will be demonstrated in further detail in the following sections.

As a summary of Table~\ref{fig:tableproai}, the most important key aspects are the performance of the CPU and GPU. ProAI outshines especially with an 8-core ARM CPU and a GPU Volta architecture with 32GB GPU memory, due to the share system memory, at a power consumption of max 30W (GPU). The computing power of ProAI exceeds the P2000 GPU of the workstation while using FP16 (half precision) by the factor 73, while using FP64 almost by 9 times. When using FP32, the workstation GPU is stronger than the ProAI, as it is apparently optimized primarily for workstation tasks like rendering. Further, the LPDDR4 memory also has a significantly higher read/write rate than the other platforms, which makes the data connection between CPU and GPU significantly faster.

Benchmark result for the ProAI and other platforms are presented for a multitask DNN. For this purpose, we selected the Multitask-CenterNet architecture, which is briefly explained in the next section.

\section{The Multitask-CenterNet (MCN) Architecture – A Brief Description}
\label{sec:mcenternet}

In most cases, multitask networks are built on a common backbone, in order to extract basic features. Branches for the various tasks, which can consist of several layers, emerge from the backbone. The selection of the emerging point can prove difficult, because in order to achieve the best possible result, this point is different for every task. The multitude of possible combinations to solve such problems is shown in \cite{surveymultitask2020}, where different approaches are presented.

\begin{figure}[t]
\begin{center}
\includegraphics[width=0.5\textwidth, trim= 125 75 135 75,clip]{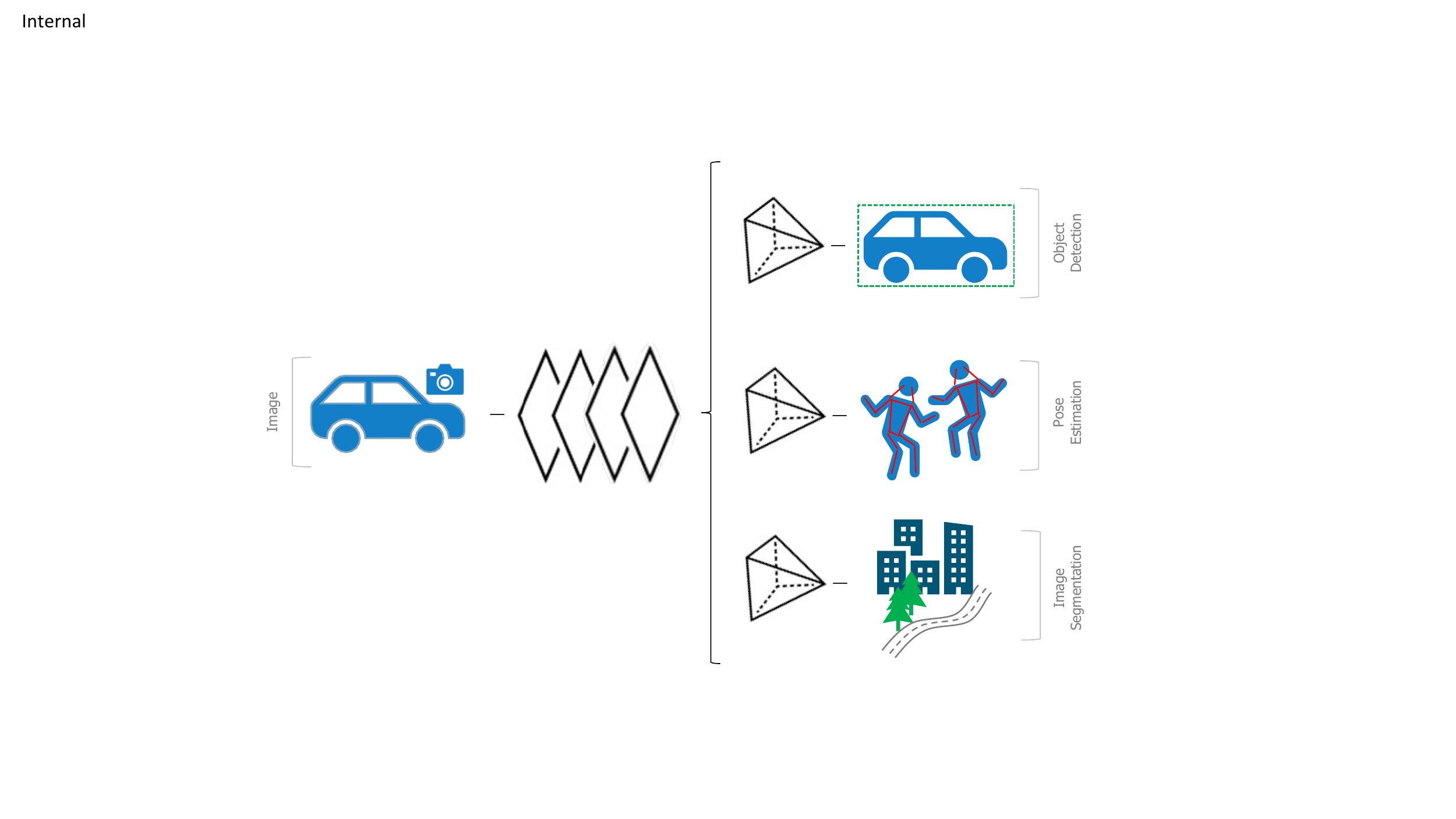}
\end{center}
   \caption{Architecture of Multitask-CenterNet (MCN) \cite{efficientmultitask2021}. The network has a single image input and a shared backbone. On top are separate prediction heads, one for each task. In this case, Object Detection (Det), Semantic Segmentation (Seg) and Pose Estimation (Pos) are shown.}
\label{fig:mcenternet}
\end{figure}

The Multitask-CenterNet (MCN) \cite{efficientmultitask2021} is based on the CenterNet architecture \cite{objectsaspoints2019}. The MCN uses a common backbone like a ResNet50 or ResNet101 \cite{residualimagerecognition2015} and several different prediction heads for perception related tasks. An illustration of the MCN architecture is show in Figure~\ref{fig:mcenternet}. 

To evaluate the performance of the MCN architecture on SBCs, we used three different heads: object detection (Det), semantic segmentation (Seg) and human pose estimation (Pos). The advantage of the MCN architecture is very clear; while the necessary expensive computational operations in the backbone only need to be performed once, each prediction head acts independently. Moreover, each head is optimized for its own respective task. Especially in the field of autonomous driving, the possibility of using these tasks concurrently is required to design a robust and versatile agent. This network forms an ideal benchmark basis, as it is both state-of-the-art and directly related to the automotive domain. Using single tasks, semantic segmentation reached 49\% mIoU and detection achieved 36.7\% mAP. With the combination of segmentation and detection, the MCN achieved an accuracy of 48.9\% mIoU (Seg) and 36.7\% mAP (Det), which is an improvement compared to using the tasks separately. When adding the third task, human pose estimation, the accuracy of both segmentation and detection tasks drops to 41.7\% mIoU (Seg) and 25.1\% mAP (Det). This is most likely due a class imbalance of the setup, where the COCO-dataset has a total of 80 classes where only one of the classes contains pose-estimation labels (class: "person"). The training of non-human classes for detection and segmentation is at least one reason for the loss of accuracy. A detailed explanation with further details of the MCN architecture and its results for different setups are presented in the work \cite{efficientmultitask2021}. In the rest of the paper prediction accuracy is not considered in the benchmark, as it is independent of the platform. The next section will define the dataset, metrics and benchmark method that are used in this paper.

\section{Dataset, Metrics and Benchmark Method}
\label{sec:dataset}
In this section the dataset and metrics used in this benchmark are presented before explaining the process and content of the benchmark itself.

\subsection{Dataset and Metrics}
\label{subsec:dm}
When selecting the dataset, various aspects had to be taken into account. Regarding the area of application, a data set from the automotive context is important. However, this harbors some difficulties with regard to the tasks used within the Multitask-CenterNet (detection, segmentation, human pose estimation). Many datasets often use common label types like semantic segmentation or bounding boxes but missing out on human pose labels, where data fusion is inevitable if the goal is to train a multitask network. This represents a challenge of its own, especially in terms of accuracy, which is why this is also a separate research area in combination with domain adaptation in the field of deep learning. Furthermore, the goal of this benchmark is the comparability of different hardware platforms to the ProAI under as similar conditions as possible. In addition, we find similar general conditions are more important for performance-oriented benchmarks with regard to the data than the actual context of the data.

The aforementioned points are ideally combined in the COCO dataset \cite{coco2016}. The dataset provides a wide variety of labels for object detection, semantic segmentation and human pose estimation and is commonly used with a variety of different DNN and publications, which is the ideal basis for this benchmark. 

The most essential part for mobile applications is efficiency and the amount of throughput per second of the hardware. Therefore, the metrics frames-per-second (FPS), inference time, memory usage and energy efficiency (W) (W$_{total}$ and W/fps) were used for the benchmark. For the benchmark evaluation, we use the specifications given in Table \ref{fig:tableproai} as the maximum GPU power consumption. 

\subsection{Benchmark Method}
\label{subsec:bm}

\begin{algorithm}
\caption{DataLogger \newline Custom software to measure relevant system data for inferencing models. Reading sensor data and adding it to a structure which directly writes it to a file.}
\label{alg:algorithm1}
\begin{algorithmic}  
\WHILE{model inferencing}
    \STATE \textbf{init} data\_struct 
    \STATE data\_struct $\leftarrow{}$ cpu utilization
    \STATE data\_struct $\leftarrow{}$ memory utilization
    \STATE data\_struct $\leftarrow{}$ system temperature 
    \IF{OS == linux}
    \STATE gpu utilization = tegrastats()
    \ELSE 
    \STATE gpu utilization = nvidia-smi()
    \ENDIF
    \STATE data\_struct $\leftarrow{}$ gpu utilization
    \STATE print(data\_struct)
    \STATE log\_to\_csv(data\_struct)
\ENDWHILE
\end{algorithmic}
\end{algorithm}

For the benchmark of SBCs, the NVIDIA Jetson Nano and ZF ProAI are used against a modern workstation laptop, which can be regarded as a stationary platform with a total of 230W of total power. However, it must be taken into account that the mobile platforms have ARM-based CPUs, whereas the workstation laptop has an x64\_x84 CPU architecture, which means that the results of pre- and post-processing (in particular the FPS and inference time) can only be compared to a limited extent. The laptop uses an Intel i7-8850H as CPU and a NVIDIA P2000 as GPU. First two different networks were compared with the Multitask-CenterNet: the SSD300 \cite{ssd1015} for object detection and the DeeplabV3 \cite{rethinkingatrousconv2017} for semantic segmentation, both with a ResNet50 Backbone. These two network architectures are representative networks of their areas of use and are frequently used in the literature. We used standard implementations of each network without further improvement regarding network inference or pre- and post-processing steps. Additionally, in order to see the influence of the different tasks of the Multitask-CenterNet, several combinations of them were used during the benchmark: 1) Seg: only semantic segmentation, 2) Det: only object detection (Det), 3) Det + Pos: both object detection and human pose estimation together, and 4) Seg + Det + Pos: all heads combined together. The necessary arithmetic operations are directly related to the complexity of the network, which is different for the tasks presented, especially semantic segmentation has a high computational requirement.
The utilization of the CPU and GPU can therefore provide information about how much computing power is still available for other applications during the inference. 

The complete software is implemented in Python in combination with the deep learning framework Pytorch \cite{pytorch2019}. In order to use the models on the NVIDA-GPU, the frameworks CUDA and cuDNN are required. Several other libraries are used to implement the image pre- and post-processing (e.g. OpenCV, Numpy). Since the main task of SBCs for deep learning in the automotive area is mostly to inference models and not the training process itself, we only measured timings and utilization during the inference process while using pretrained weights. 
The pretrained weights for the Multitask-CenterNet were provided by us, the other weights were provided by the Pytorch framework itself. As inferencing data, the test set “COCO test2017” is used, which includes 41k images. For our benchmark, only 5k images of the test dataset is used for the benchmark itself. This is a sufficient amount of data to test memory and inference timings of each model. Custom software is used to record the test metrics. All necessary system values are queried with a frequency of 0.5Hz and displayed live in a terminal as well as logged to a file, in order to be able to use those values afterwards. Since ProAI and Jetson Nano have ARM-based systems with a linux operating system, whereas the workstation PC uses windows as operating system, the code slightly differs due to specific terms of the operating system (mostly affects how to read the GPU utilization). A simplified process of the measurement is shown in Algorithm \ref{alg:algorithm1}.

\begin{figure}[t]
\begin{center}
\includegraphics[width=0.5\textwidth,clip, trim=5 10 5 5]{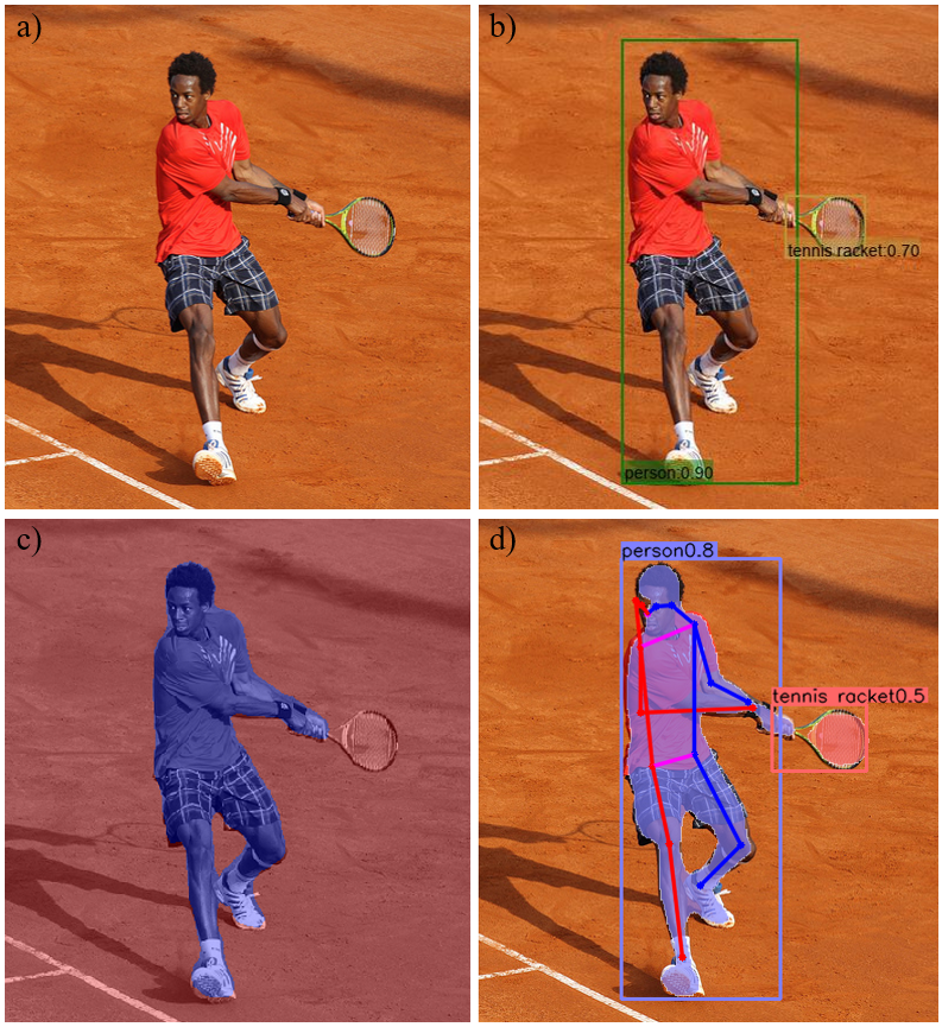}
\end{center}
   \caption{Qualitvative results of different models. a) original image b) result of Single Shot Detector (SSD) with ResNet backbone c) result of DeeplabV3 (the tennis racket is not segmented because the model was pretrained on only 21 classes of COCO Dataset which does not include the class tennis racket) d) result of the Multitask-CenterNet (MCN).}
\label{fig:qualitative}
\end{figure}

\section{Benchmark Results and Discussion}
\label{sec:benchmark}

\begin{figure}[t]
\begin{center}
\includegraphics[width=0.48\textwidth,clip,trim=15 22 10 15]{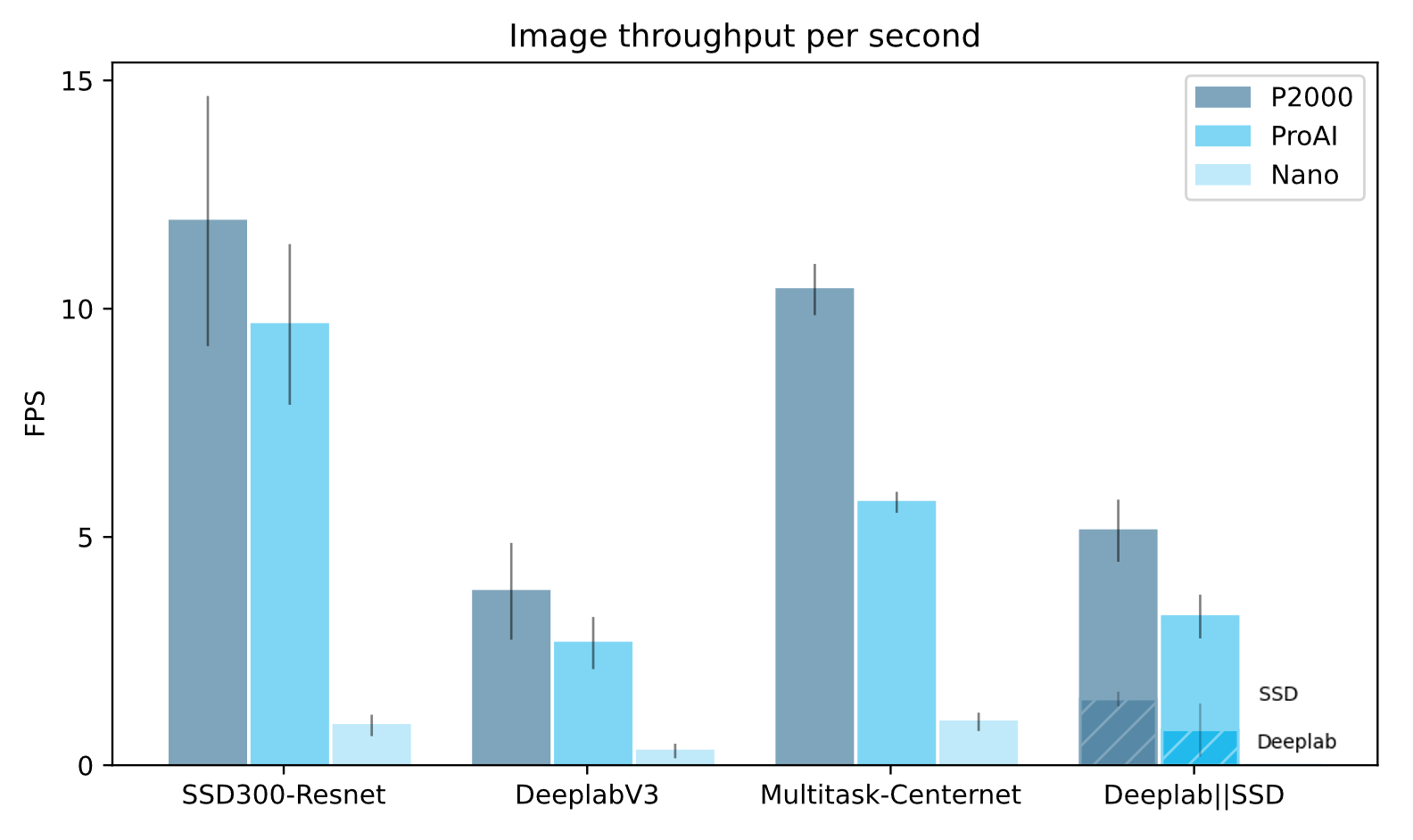}
\end{center}
   \caption{Comparison of different networks on all platforms. The Results show inference with pre- and postprocessing. The most right result shows the inference when SSD and Deeplab are running simultaneously on P2000 and ProAI (Jetson Nano board could not run both models simultaneously, due to insufficient amount of memory).}
\label{fig:quantitative_1}
\end{figure}

\begin{figure*}[h]
\begin{center}
\includegraphics[width=1.0\textwidth]{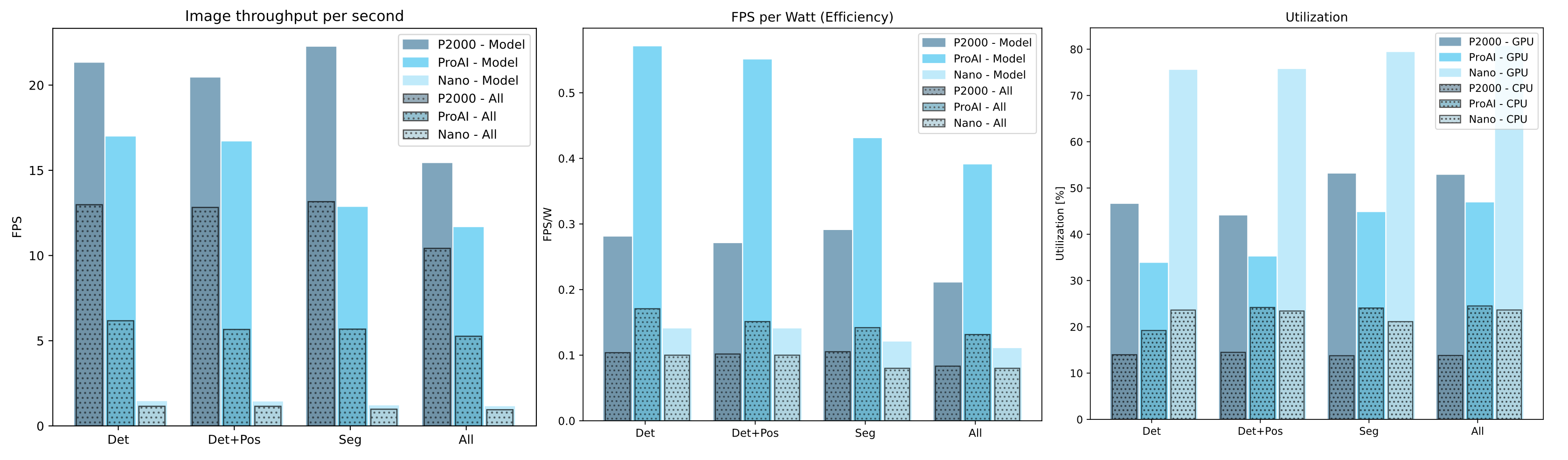}
\end{center}
   \caption{Benchmark Results: Left: The FPS for each task and platform is shown, distinguished between model performance (monochrome) and complete process performance (dotted). Middle: Shows the efficiency with FPS per watt and the same distinction as in the left graph. Right: Shows the GPU (monochrome) and CPU (dotted) utilization for each task.}
\label{fig:quantitative_2}
\end{figure*}

In this section the performance evaluation and comparison of each platform is presented, including ZF ProAI, the NVIDIA Jetson Nano and a workstation laptop with a P2000 GPU.\\
\newline
First the qualitative results of the Multitask-CenterNet (detection, semantic segmentation and human pose estimation) compared to state of the art networks (SSD300 for object detection and DeeplabV3 for semantic segmentation) are shown in Figure~\ref{fig:qualitative}. Since the networks used for this comparison were only pretrained on 21 classes of the COCO dataset\footnote{Pretrained weights available on \url{www.pytorch.org}} an image with the class "person" was chosen to get a fair comparison, as it was available in each of the training processes of all networks.

We have split the quantitative benchmark results into two parts: performance of the MCV against state of the art models (SSD and DeepLab) on different platforms, and performance of different MCN settings on different platforms. A detailed comparison of the MCN network against SSD and DeepLab for different platform results are presented in Figure~\ref{fig:quantitative_1}. In all measurements, both the inference time of the network itself and the pre- and post-processing of the results were considered. Additionally, to enable a realistic statement, a batch size of one was used during inference, as with the application of a (single) camera. The DeeplabV3 performance was the worst on all platforms which was not very surprising due to the computationally intensive segmentation task. The SSD, on the other hand, had the highest throughput on the workstation (12.46 FPS) and the ProAI (9.85) as expected. But on the Jetson Nano the Multitask-CenterNet was with 0.95 FPS slightly faster than the SSD with 0.87 FPS. This was quite surprising, due to the fact, that the Multitask-CenterNet combine detection and semantic segmentation in this particular case. In order to create a more realistic comparison to the Multitask-CenterNet, the inference of the SSD and DeeplabV3 was run simultaneously (to have two different task running simultaneously). This scenario clearly shows the advantage of the Multitask-CenterNet, especially on platforms with very limited resources. The Jetson Nano board could not run both models simultaneously, due to an insufficient amount of memory, but the Multitask-CenterNet with very promising results. Also the ProAI achieves much better results with the Multitask-CenterNet than with SSD and Deeplab in parallel. When comparing the workstation and the ProAI when running those two models in parallel, the difference between ProAI and workstation is even smaller in comparison to a single model inference.  

The detailed results regarding the different settings of Multitask-CenterNet for different platforms are shown in Figure~\ref{fig:quantitative_2}. FPS and FPS/Watt are distinguished between the performance of the model prediction itself, in which only the GPU computing part is very high, and the whole process including image loading, network prediction and pre-/post-processing. The frames-per-second, which are derived from the inference time, are best on the workstation laptop with the Quadro P2000 GPU, as expected. The Jetson Nano has a significantly slower prediction time using all task combined, which makes the differences between detection only and all tasks at the same time hardly recognizable (Figure~\ref{fig:quantitative_2} – left). Despite the big differences in computing power, ProAI only has 3.5 FPS less in total compared to the workstation laptop. 

Considering the efficiency, the ratio between FPS and power consumption of the respective platform, ProAI is ahead by a clear margin (Figure~\ref{fig:quantitative_2} – middle). For this relation, the FPS derived by the inference time and the maximum GPU consumptions from Table ~\ref{fig:tableproai} were used. The ProAI achieves almost twice as much FPS/Watt in all task combinations in the analysis except segmentation (Seg). The reason for the backlog in the area of segmentation is mainly found in the post-processing step, since the complex calculation of the predicted heatmaps and up-sampling steps is CPU based until now. Therefore, there is an advantage for the power station with the significantly more powerful CPU.

Another important part is the utilization of CPU and GPU of the systems. Striking is the similar CPU load of the two ARM-based boards in contrast to the x64\_86 CPU of the workstation laptop (Figure~\ref{fig:quantitative_2} – right). In addition to the significantly faster post-processing on the workstation laptop, this additionally shows the greater computing power of the Intel CPU over the ARM processor. However, it must be considered that the Intel CPU has a max. consumption of 45 W, whereas ARM processors only use a few watts in total. However, in contrast to the other two platforms, ProAI has a significantly lower GPU utilization, despite the lower number of cores compared to the P2000. The high utilization of the GPU with three tasks on the Jetson Nano board shows that it has nearly reached its limit and has not much potential left for more than three tasks with the Multitask-CenterNet. With a maximum GPU utilization of 46\% on ProAI, there is still a lot of free potential which can be used for further tasks or the parallel calculation of other networks.

\section{Conclusion}
\label{sec:conclusion}
The requirements for single board computers in ADAS are enormously high, both in terms of the computationally expensive and complex applications and in terms of the necessity of quality and safety requirements. ZF shows with the development of its ProAI that such an SBC can meet all necessary requirements for example running power intensive multitask deep learning models. Using the Multitask-CenterNet (MCN) architecture, the performance of ProAI is compared to another SBC as well as a stationary platform. This architecture enables significantly better performance on platforms that are particularly limited in terms of resources, as is often the case with edge devices. ZF ProAI demonstrated with a clear lead that high efficiency and performance can be combined in a single SBC. Nevertheless, the utilization results of the ARM processor have shown that there is further potential for improvement, especially in the area of pre- and post-processing. Furthermore, the innovation cycles in high performance CPUs and GPUs are very short. Today we already see the next generations coming up which have again enormous increase of efficiency and performance. This will enable even more complex algorithms by lower power consumption.

\section*{Acknowledgment}
This work is in part funded by the German Federal Ministry for Economic Affairs and Energy (BMWi) through the grant 19A19013Q, project "KI Delta Learning".

\section{References}
\label{sec:ref}

{\small
\bibliographystyle{ieee_fullname}
\bibliography{egbib}
}

\end{document}